\documentclass{llncs}

\usepackage{hyperref}
\usepackage{url}
\usepackage{xcolor}
\usepackage{algorithmic}
\usepackage{bbm}
\usepackage{multirow}
\usepackage{nccmath}
\usepackage{algorithm, algorithmic}
\usepackage{amsfonts}
\usepackage{amssymb}
\usepackage{amsmath}
\usepackage{bbm}
\usepackage{color}
\usepackage{floatflt}
\usepackage[T1]{fontenc}
\usepackage{graphicx}
\usepackage{hyperref}
\usepackage{latexsym}
\usepackage{multirow}
\usepackage[pdftex]{pstricks}
\usepackage{tabularx}
\usepackage{times}
\usepackage{url} 
\usepackage[normalem]{ulem} 
\usepackage{verbatim}
\usepackage{wasysym}
\usepackage[normalem]{ulem}
\usepackage{subfigure}

\begin{document}

\title{Constructing Variables Using Classifiers as an Aid to Regression: An Empirical Assessment}
\titlerunning{Constructing Variables Using Classifiers as an Aid to Regression: An Empirical Assessment}

\author{Colin Troisemaine \and Vincent Lemaire }

\authorrunning{Colin Troisemaine, Vincent Lemaire}

\institute{Orange Innovation, Lannion, France}

\maketitle

\begin{abstract}
This paper proposes a method for the automatic creation of variables (in the case of regression) that complement the information contained in the initial input vector.
The method works as a pre-processing step in which the continuous values of the variable to be regressed are discretized into a set of intervals which are then used to define value thresholds.
Then classifiers are trained to predict whether the value to be regressed is less than or equal to each of these thresholds. The different outputs of the classifiers are then
concatenated in the form of an additional vector of variables that enriches the initial vector of the regression problem. The implemented system can thus be considered as a generic pre-processing tool.
We tested the proposed enrichment method with 5 types of regressors and evaluated it in 33 regression datasets.  Our experimental results confirm the interest of the approach.
\end{abstract}

\section{Introduction}

Learning techniques can be divided into two main categories according to their main purpose: those used to describe data (descriptive methods) and those used to predict a (more or less) observable phenomenon (predictive methods).
Predictive methods use a set of labeled data to predict and explain one or more (more or less) observable phenomena.

In the case of regression, this involves predicting the value of a numerical variable (noted $y$), for example the amount of an invoice, using a set of explanatory variables (a vector noted $X$).
In the case of machine learning, we seek to learn a function $f$ such that $y=f(X)$ using a machine learning algorithm and a training set, a set of $N$ input-output pairs $(X_i,y_i), i={1, ..., N}$.
During this modeling phase, it is often necessary to create new variables that better describe the problem and enable the model to achieve better performance.
This is what we call the "engineering process of creating new explanatory variables" \cite{sondhi2009feature}.
The idea here is here that the new variables (a vector, in this case $X'$) will provide additional information.
By automating the generation of these "new variables", more useful and meaningful information can be extracted from the data, in a framework that can be applied to any problem. This allows the machine learning engineer to devote more time on more useful tasks.

The aim of the article presented here is to propose a method for automatically creating variables (in the case of regression) that complement the information contained in the vector $X$ in order to predict the values of the dependent variable $y$. The proposed method first transforms the regression problem into several classification sub-problems, then integrates the results in the form of additive variables ($X'$). The 'augmented' vector, $X''=X \cup X'$, is then fed into standard regressors to measure the contribution of the created variables. The advantages of this approach are presented in the form of a detailed experimental study.

\section{Proposal}

The proposed method relies on transforming the regression problem into a classification problem to construct the additional variables. This idea has been proposed before, but our approach differs from the state of the art, as explained below in section \ref{sota}. Having presented this difference, section \ref{methode} describes the proposed method in detail and section \ref{implementation} describes the choices made for its implementation.

\subsection{Related work}
\label{sota}

Solving a regression problem using classification models is an approach that has already been explored. This process has been described in numerous papers \cite{Ahmad2012NovelEM,ahmad2018learning,TUD-KE-2010-01,jf:IJCAI-11,8852133,TORGO1997} and generally consists of two main steps: (i) discretization (also called partitioning or binning) of the target variable to enable the use of classifiers on the dataset; (ii) prediction of the regression is then usually performed by calculating the mean (\cite{Ahmad2012NovelEM,ahmad2018learning,TUD-KE-2010-01,8852133}) or median (\cite{ahmad2018learning,jf:IJCAI-11,TORGO1997}) of the instances within the fragment of discretized output predicted by the classifier.

The method we propose and present below differs from these works in that the classifiers used by the method are solely aimed at adding complementary explanatory variables to the initial explanatory variables (native variables). The augmented vector is then positioned at the input of a standard regressor. This regressor directly predicts the target variable without any further transformation or estimation operations.
As the next section will show, the proposed method is linked to a conditional estimate of the density function of $y$. It would then be interesting to test other, less computationally-intensive methods of variable creation in the same framework \cite{rothfuss2019conditional,Holmes:2012,tutz2021ordinal,langford2012predicting} in future work. But the general principle remains the same.

\subsection{General principle}
\label{methode}

\begin{figure}[h]
	\begin{center}
		\includegraphics[width=0.45\textwidth]{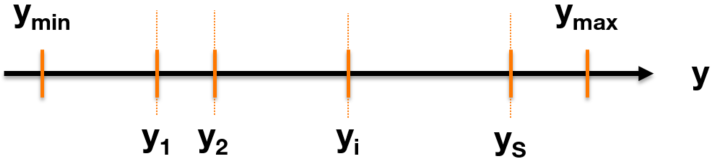}
		\caption{Example of thresholds}
		\label{fig:seuils}
	\end{center}
\end{figure}

The idea is to take opportunistic advantage of the progress made in recent years by classifiers in the literature. The principle of the first step of the proposed method is to transform the regression problem into one (or more) classification problem(s) by \textit{discretizing} the space of variation of the variable to be regressed. This step consists in defining thresholds ($S$) on the space of the variable to be regressed (see figure \ref{fig:seuils}).

These thresholds will be defined using the values of the variable to be regressed in the training set. They will be used to define membership classes ($C=\{C_1, ..., C_i, ..., C_S\})$. Examples include classes defined on the basis of value inferiority thresholds ($C_i := \mathbb{1}_{y \leq y_i}$) or classes defined on the basis of membership of value intervals ($C_i := \mathbb{1}_{y \in ]y_i,y_{i+1}]}$), etc.

Once the classes have been encoded and one (or more) classifiers have been trained (using the training set), it is then possible to predict the membership of individuals in the previously defined classes. The predictions of the classifier(s) on each individual will then be used to create a new "extended" data set, either for the training set or the test set.

\begin{figure}[h]
	\begin{center}
		\includegraphics[width=0.75\textwidth]{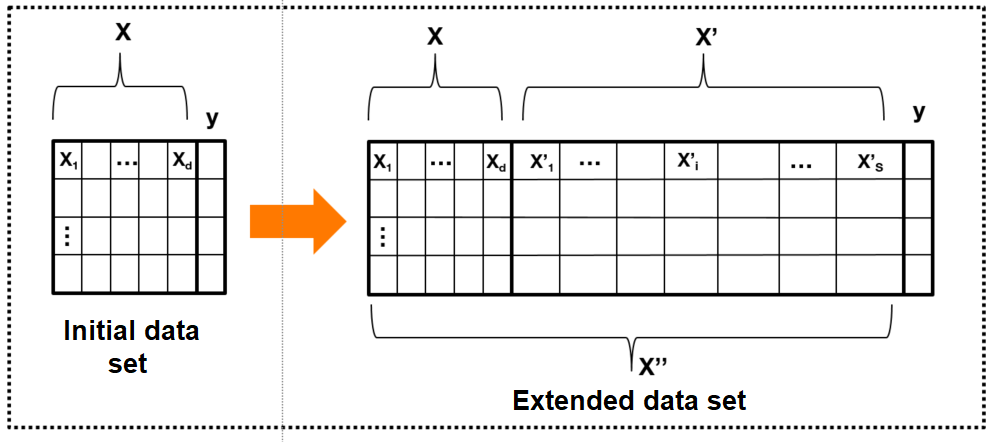}
		\caption{Data set extension.}
		\label{fig:DatasetExtension}
	\end{center}
\end{figure}

The figure \ref{fig:DatasetExtension} illustrates the extension of the vector $X$ (with $d$ components) to the vector $X''$ (with $d+S$ components). The section \ref{implementation} will describe in more detail what the vector $X'=\{X'_1, ..., X'_i, ..., X'_S\}$ from the prediction of the classifier(s) is composed of.

Once the $X$ vector has been "extended", the regression model can be trained and predicted on the new data. We can then compare the regressor's performance when trained on the original data set (using only $X$) with its performance when using the extended set ($X''=X \cup X'$). The assumption being that the regression model will have a better intuition of the general position of individuals in the space of the variable to be regressed and will produce better results.

\begin{figure}[H]
	\begin{center}
		\includegraphics[width=0.75\textwidth]{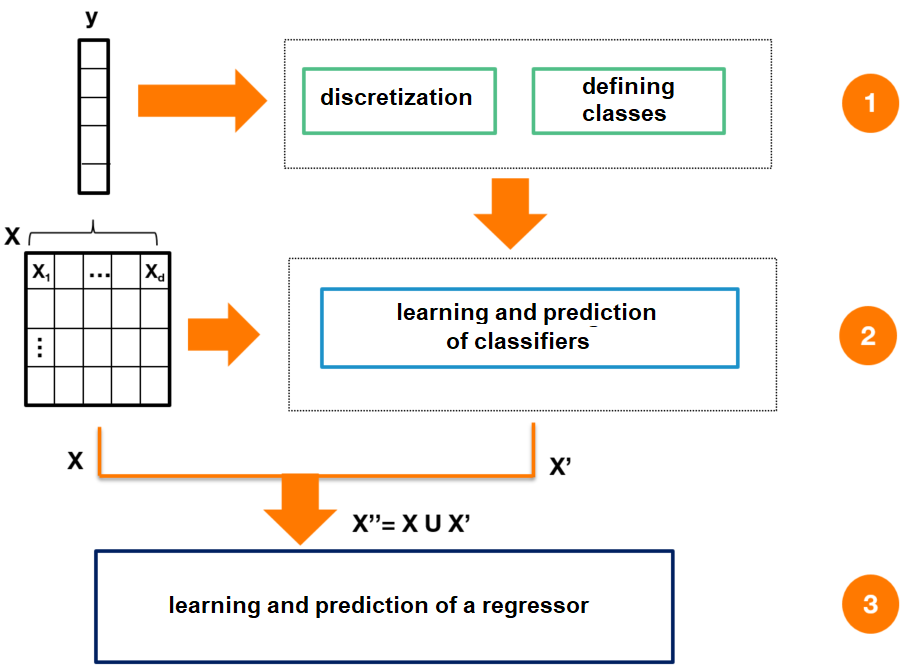}
		\caption{Diagram of the general principle of the method.}
		\label{fig:generalprinciple}
	\end{center}

\end{figure}

Figure \ref{fig:generalprinciple} summarizes the process followed by the proposed method: (i) the first step is to transform the regression problem into a classification problem. To do this, the target variable $y$ is first discretized (see section \ref{implementation}), then classes are defined using the thresholds thus defined. (ii) In a second step, the classifiers are trained using the original descriptive variables $X$ and the new classes derived from $y$. Then the prediction, of the classifiers using the initial vector $X$, is used to extract new features, i.e. $X'$. (iii) Finally, the regression model can be trained using the vector $X''=X\cup X'$.
The proposed method is thus based on a discretization and a class encoding mechanism. These two processes can be implemented in different ways. An example of implementation is given below in the section \ref{implementation}.

\subsection{Implementation}
\label{implementation}

One of the most important aspects of the proposed method is the association of classes with instances. These classes will be used to train the classifiers. This process is based on two steps: the definition of thresholds and the encoding of classes. For the definition of {\it threshold placement}, which amounts to an unsupervised discretization of the variable to be regressed, there are numerous possibilities in the literature, such as ``EqualWidth'', ``EqualFreq'', etc. For {\it class encoding}, it is possible to formulate the problem either as an $S$ class classification problem (where $S$ denotes the number of thresholds), or as $S$ binary classification problems, or as a multi-label classification problem.

The choice of discretization method and the number of thresholds are linked, and this choice represents a trade-off  between (i) classification efficiency and (ii) information input for the initial regression problem. In the case of a single classifier predicting the membership interval of individuals, it seems obvious that the larger S is, the more precise the information provided to the regressor will be, but the more difficult the problem will become to learn. On the other hand, if we formulate the problem as $S$ binary classification problems, each classifier will have a problem of the same difficulty to solve, regardless of the value of $S$. Preliminary tests carried out on the set of datasets described in Section \ref{datasets} have confirmed this behavior. The final choices for the various elements associated with the method proposed in this article are described below.

For the {\bf discretization method} - The ``Equal Frequency" method was chosen because, unlike the ``Equal Width'' method, it does not run the risk of creating intervals that contain no individuals. It also ensures that there are no classification problems where the minority class (extreme threshold on the left, see Figure \ref{fig:seuils}) represents less than $\frac{1}{S+1}$ percent of the individuals.\\

For the nature of the {\bf classification problem} - Preliminary tests, posing the problem as $S$ binary classification problems gave better performance, both in learning and in deployment. The results are also more robust (better generalization), which is important if we want the added variables to benefit the regressor. The classifiers used in the rest of this article are Scikit-Learn's Random Forests \cite{RandomForestsBreiman} with 100 trees (but any other powerful and robust classifier could be used) with their default parameters.\\

For the {\bf number of thresholds} - The first intuition (confirmed by the experiments) is that the greater the number of thresholds defined, the greater the potential performance gain. Indeed, the more thresholds defined, the finer the discretization and the closer the classifiers' prediction will be to the true regression value. In the experimental phase of this article, we will illustrate the performance of our method as a function of the number of thresholds defined. The number of thresholds defined on the space of the variable to be regressed is therefore a hyper-parameter of our method.\\

For the definition of {\bf classes associated with thresholds} - We chose classes associated with  value inferiority thresholds ($C_i := \mathbb{1}_{y \leq y_i}$) (see Section \ref{methode} and Figure \ref{fig:seuils}).\\

For the {\bf variables extracted} from the classifiers - Each classifier will be trained to predict whether the data provided is below or above the threshold at which it is assigned a class. As the method is intended to be generic, it was decided to extract the conditional probabilities predicted by each classifier. Indeed, this information is available for the vast majority of classifiers in the literature. Conditional probabilities\footnote{Note: as the selected classifiers are random forests, the conditional probability corresponds to the proportion of trees that voted for the class in question.} of class 1 (i.e. if $y \leq y_i$) predicted by each classifier will therefore make up the vector {\footnotesize $X'=\{X'_1, . ..., X'_i, ..., X'_S\}=\{P(C_1=1|X), ..., P(C_i=1|X), ..., P(C_S=1|X)\}$}; $X'$ will then be added to the initial data vector ($X$).

\section{Experimental protocol}

\subsection*{Code availability}
The code for experiments is available at the following url:
{\url{https://github.com/ColinTr/ClassificationForRegression}}

\subsection{Data sets and pre-processing}
\label{datasets}

To carry out the analysis of the proposed method, we selected a large collection of regression datasets from the UCI Repository \cite{Dua:2019} or from Kaggle \cite{kaggle}. 
A total of 33 datasets were used, of which 23 consisted of more than 10,000 individuals. The remaining 10 datasets range from 1,030 to 9,568 individuals (see Table \ref{table:DatasetsDescription}). This selection was largely influenced by the article by M. Fernandez-Delgado et al. \cite{FERNANDEZDELGADO2019}. The selected datasets were, according to the categorization of Fernandez-Delgado et al. "{\it large and difficult}". 
Some of these datasets included several subsets of data with different target variables to be regressed, and therefore different regression problems. It was decided not to include multiple regression problems from the same datasets in order to avoid giving more weight to certain datasets and creating a bias in the comparison of methods that might favour one method over another. The reader can refer to the "Dataset name" column of the table \ref{table:DatasetsDescription}, which identifies the regression problem used for each dataset.

{\bf Pre-processing:} As in \cite{FERNANDEZDELGADO2019}, two pre-processings were carried out before entering the process described in Figure \ref{fig:generalprinciple} (see Section \ref{methode}): (i) recoding of categorical variables using full disjunctive coding; (ii) deletion of date variables, constant variables, individual identifiers, collinear variables and other variables potentially to be 'regressed'. Three further pre-processings were performed on each learning fold. These are, in order of completion: (i) normalization of numerical variables (centering - reduction); (ii) normalization of the variable to be regressed (centering - reduction) then transformation using Box-Cox \cite{Atkinson2020TheBT}; (iii) creation of thresholds for class definition as described above. For each of these three pre-processings, the associated statistics were calculated on the training set and then applied to the training and test sets. In the following results presentation, the RMSE results (see metric used below) will be given without performing the inverse Box-Cox transformation. Finally, examples with missing values have been removed from the initial dataset. Dataset statistics after pre-processing are presented in Table \ref{table:DatasetsDescription}.

{\bf Train-Test split and model optimization:}
Each dataset has been split into a "10-fold cross validation", resulting in 10 training and test sets. For the three models requiring parameter optimization (AR, FA and XGB see Section \ref{regresseur}): 30\% of the training set was reserved for optimizing the model parameters in a "grid-search" process, often avoiding over-training. 
Then, provided with the "right" training parameters, the model was trained using the remaining 70\% of the training set. Finally, once the model had been trained for this fold, its performance was measured. 
This rigorous process was carried out for all datasets, all folds and all $S$ values, resulting in the training of thousands of models, but allowing a rigorous test of the proposed method.

\begin{table*}[!h]
    \centering
    \fontsize{6}{8}\selectfont
    \setlength{\tabcolsep}{2pt}
    \begin{tabular}{|l|l|l|l||l|l|l|l|}
        \hline
        Data set       & \#individuals    & \# variables & target & Data set       & \#individuals    & \# variables & target\\ \hline
        \hline
        3Droad               & 434,874        & 3       & altitude   & geo-lat              & 1,059          & 116     & latitude\\ \hline
        air-quality-CO       & 1,230          & 8       & PT08.S1(CO)  & greenhouse-net       & 955,167        & 15      & synthetic var  \\ \hline
        airfoil              & 1,503          & 5       & scaled sound  & KEGG-reaction        & 65,554         & 27      & edge count\\ \hline
        appliances-energy    & 19,735         & 26      & appliances    &   KEGG-relation        & 54,413         & 22      & clustering coef  \\ \hline
        beijing-pm25         & 41,758         & 14      & PM2.5 & online-news          & 39,644         & 59      & shares \\ \hline
        temp-forecast-bias   & 7,752          & 22      & Next\_Tmax &  video-transcode      & 68,784         & 26      & utime\\ \hline
        bike-hour            & 17,379         & 17      & count  &  pm25-chengdu-us      & 27,368         & 20      & PM\_US Post\\ \hline
        blog-feedback        & 60,021         & 18      & target  &  park-total-UPDRS     & 5,875          & 16      & total\_UPDRS\\ \hline
        buzz-twitter         & 583,250        & 70      & discussions &  physico-protein      & 45,730         & 9       & RMSD \\ \hline
        combined-cycle       & 9,568          & 4       & PE   & production-quality   & 29,184         & 17      & quality \\ \hline
        com-crime            & 1,994          & 122     & Violent crimes &  CT-slices            & 53,500         & 384     & reference  \\ \hline
        com-crime-unnorm     & 2,215          & 134     & Violent crimes  & gpu-kernel-perf      & 241,600        & 14      & Run1 (ms) \\ \hline
        compress-stren       & 1,030          & 8       & compressive strength & SML2010              & 4,137          & 26      & indoor temp \\ \hline
        cond-turbine         & 11,934         & 15      & gt turbine & seoul-bike-sharing   & 8,760          & 9      & Rented Bike Count  \\ \hline
        cuff-less            & 61,000         & 2       & APB  & uber-location-price  & 205,670        & 5       & fare amount\\ \hline
        electrical-grid-stab & 10,000         & 12      & stab  &  year-prediction      & 515,345        & 90      & Year \\ \hline
        facebook-comment     & 40,949         & 53      & target & & & & \\ \hline
	\end{tabular}
    \caption{Description of the 33 data sets. The first column indicates the name of the dataset (or sub-dataset) from UCI or Kaggle.  The second and third columns give the initial number of individuals and (explanatory) variables in the dataset (after pre-processing). Finally, the fourth column indicates the name of the target variable to be regressed.}
	\label{table:DatasetsDescription}
\end{table*}

\subsection{Metric used for results}

When comparing regression models, the aim is to find a model without over- or under-fitting that achieves a low generalization error, which characterizes its predictive performance. Various metrics are proposed in the literature, such as the root mean square error (RMSE), the mean absolute error (MAE), the R-Square, etc... There seems to be no consensus on the "best" metric to use, although some papers do offer comparisons \cite{BotchkarevMetrics}. In the rest of the results, RMSE has been chosen and is defined as:
{\small $RMSE = \sqrt{\frac{1}{n}\sum_{i=1}^{n}(y_i-\hat{y}_i)^2}$}, where $n$ is the number of individuals, $y_i$ the desired value and $\hat{y}_i$ the output of a regressor.

\subsection{Tested Regressors}
\label{regresseur}

Five regressors were used in the experiments, using different frameworks. They are briefly described below, although they are well known in the machine learning community.\\

{$\bullet$ Linear Regression (LR): Regression refers to the process of estimating a continuous numerical variable using other variables that are correlated with it. This means that regression models take the form of $y=f(X)$ where $y$ can take on an infinite number of values. In this article, we have chosen a multivariate linear regression whose model, $f$, is determined by the method of least squares. We have used Scikit-Learn version 0.24.2. This model does not require parameter optimization.\\

{$\bullet$ Regression Tree (DT): }
Decision trees are used to predict a real quantity which, in the case of regression, is a numerical value. Algorithms for building decision trees are usually constructed by dividing the tree from the root to the leaves, choosing at each step an input variable that achieves the best partitioning  of the set of individuals. 
In the case of regression trees, the aim is to maximize the inter-class variance (i.e. to have subsets whose values of the target variable are as widely spread as possible). In this article, we used Scikit-Learn version 0.24.2. This model requires parameter optimization.\\

{$\bullet$ Random Forest (RF): } Decision-tree forests (or Random Forest (RF) regressors) were first proposed by Ho in 1995  \cite{RandomDecisionForests} and then extended by Leo Breiman and Adele Cutler \cite{RandomForestsBreiman}. This algorithm combines the concepts of random subspaces and bagging. The decision tree forest algorithm performs training on multiple decision trees trained on slightly different subsets of data. In this article, we have used Scikit-Learn version 0.24.2. This model requires parameter optimization.\\

{$\bullet$ XGBoost (XGB): }
XGBoost \cite{xgboostpaper} is a boosting method. It sequentially combines weak learners that would individually perform poorly to improve the prediction of the full algorithm. A weak learner is a regressor with poor regression performance. In this boosting algorithm, high weights will be associated with weak learners having good accuracy, and conversely, lower weights with weak learners having poor accuracy. In the training phase, high weights are associated with data that has been poorly "learned", so that the next weak learner in the sequence will focus more on that data. In this article, we have used XGBoost version 1.4.1. This model requires parameter optimization.\\

{$\bullet$  Selective Na\"ive Bayes (SNB): }In the context of regression, a naive Bayes regressor (NB) whose variables are weighted by weights (NBP) can be obtained in two steps.
First, for each explanatory variable, a 2D grid is created to estimate $P(X,y)$ (see for example \cite{MODLBoulleML06}). Then, in a second step, all variables are grouped together in a Forward Backward algorithm \cite{BoulleJMLR07} to estimate their informativeness in the context of a Naive Bayes regressor. At the end of the second step, the final model (to be deployed) is a Na\"ive Bayes (using the 2D discretization found in the first step) where the variables are weighted (the weights are found in the second step). In this article we have used the Weighted Na\"ive Bayes produced by Khiops library ({\url{www.khiops.org}}). This model does not require parameter optimization.

\section{Results}

{\bf $\bullet$ Illustrative results - } Figure \ref{fig:example} illustrates the behavior of the proposed method: on the left for linear regression and the "KEGG Metabolic Reaction" dataset, in the middle and on the right for the "SML 2010" dataset, respectively for the Selective Na\"ive Bayes regressor and the Random Forest.
These 3 figures are fairly representative of the results obtained, with a more or less marked decrease in the RMSE versus the value of $S$ and the regressor considered. This decrease is most pronounced at the beginning of the curve and then levels off for a less pronounced increase beyond $S=16$. The following section presents the results obtained for each data set.

\begin{figure}[!ht]
    \centering
    \includegraphics[width=1.0\linewidth]{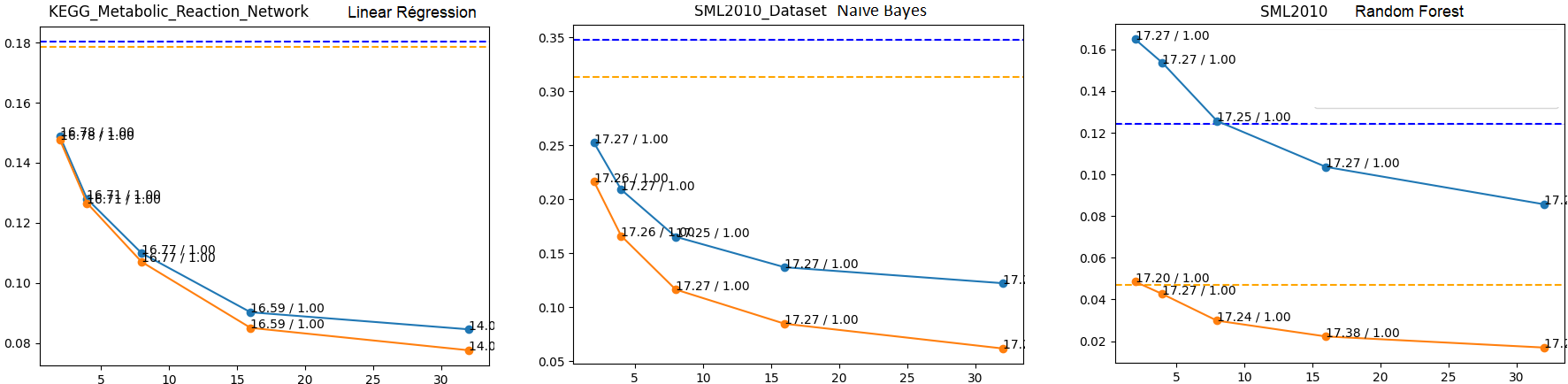}
    \caption{$S$ (horizontal axis) versus RMSE (vertical axis). The dotted  lines
    represent the initial performance of the regressor in blue for the test set and in orange for the training set. Solid lines represent the regressor's performance with the proposed method, 
    using the same color code.
    }
    \label{fig:example}
\end{figure}

{\bf $\bullet$ Results tables on all datasets - } The results obtained are presented in detail in Table \ref{table:RMSEComparison} for a number of thresholds equal to 32 ($S=32$).
For each dataset, this table gives the test results (average results over the 10 test folds (see Section \ref{datasets})) for each of the five regressors for which the addition of the vector $X'$  was tested.
In this table, a value in bold indicates a significant difference between results with the $X$ vector "Native" and the $X'$ vector "Aug" (Increased) according to paired Student's t test (p-value at 5\%). The last row of the table shows the number of losses, ties and wins for "Aug" versus "Native". We can see that the addition of the $X'$ vector essentially benefits 3 of the regressors: the linear regression, the regression tree and the naive Bayes regressor. For the random forest and XGboost, the gains and/or losses are fairly limited.

\begin{table}[!h]
    \centering
    \fontsize{8}{9}\selectfont
    \setlength{\tabcolsep}{3pt}
    \begin{tabular}{|l|c|c|c|c|c|c|c|c|c|c|}
        \hline
        & \multicolumn{2}{c|}{\textbf{Linear Regr.}} & \multicolumn{2}{c|}{\textbf{Decision Tree}} & \multicolumn{2}{c|}{\textbf{Random Forest}} & \multicolumn{2}{c|}{\textbf{XGBoost}} & \multicolumn{2}{c|}{\textbf{SNB}} \\
        \hline
        \textbf{Dataset name} & \textbf{Native} & \textbf{Aug} & \textbf{Native} & \textbf{Aug} & \textbf{Native} & \textbf{Aug} & \textbf{Native} & \textbf{Aug} &\textbf{Native} & \textbf{Aug} \\
        \hline
        3Droad               & 0,9867 & \textbf{0,0930} & 0,1213 & \textbf{0,0831} & 0,0908 & \textbf{0,0792} & 0,0935 & \textbf{0,0781} & 0,5580 & \textbf{0,0827} \\
        air-quality-CO       & 0,3269 & \textbf{0,2672} & 0,3255 & \textbf{0,2727} & 0,2667 &          0,2642 & 0,2682 &          0,2633 & 0,3428 & \textbf{0,2768} \\
        airfoil              & 0,7176 & \textbf{0,2351} & 0,4373 & \textbf{0,3194} & 0,2987 &          0,2865 & 0,2814 &          0,2752 & 0,7709 & \textbf{0,2963} \\
        appliances-energy    & 0,8260 & \textbf{0,4865} & 0,7431 & \textbf{0,5267} & 0,5456 & \textbf{0,5164} & 0,5790 & \textbf{0,5203} & 0,8354 & \textbf{0,5206} \\
        beijing-pm25         & 0,7833 & \textbf{0,4180} & 0,6013 & \textbf{0,3948} & 0,4168 & \textbf{0,3874} & 0,4149 & \textbf{0,3885} & 0,8082 & \textbf{0,3931} \\
        temp-forecast-bias   & 0,4749 & \textbf{0,2847} & 0,4668 & \textbf{0,2848} & 0,3237 & \textbf{0,2771} & 0,3085 & \textbf{0,2773} & 0,4782 & \textbf{0,2907} \\
        bike-hour            & 0,7163 & \textbf{0,2547} & 0,3294 & \textbf{0,2564} & \textbf{0,2415} & 0,2499 & \textbf{0,2208} & 0,2500 & 0,4858 & \textbf{0,2534} \\
        blog-feedback        & 0,8307 & \textbf{0,6434} & 0,6741 &          0,6694 & \textbf{0,6481} & 0,6618 & \textbf{0,6444} & 0,6652 & 0,7013 & \textbf{0,6601} \\
        buzz-twitter         & 0,7248 & \textbf{0,2183} & 0,2299 & \textbf{0,2265} & \textbf{0,2170} & 0,2200 & \textbf{0,2162} & 0,2205 & 0,2238 & \textbf{0,2191} \\
        combined-cycle       & 0,2702 & \textbf{0,1967} & 0,2647 & \textbf{0,2014} & 0,2079 &          0,1952 & 0,2041 &          0,1950 & 0,2570 & \textbf{0,1995} \\
        com-crime            & 0,6312 &          0,5485 & 0,6418 & \textbf{0,5552} & 0,5525 &          0,5505 & 0,5665 &          0,5503 & 0,5612 & 0,5523 \\
        com-crime-unnorm     & 0,6751 &          0,6031 & 0,7036 & \textbf{0,6251} & 0,6163 &          0,6157 & 0,6186 &          0,6124 & 0,6453 & \textbf{0,6340} \\
        compress-stren       & 0,6331 & \textbf{0,2732} & 0,4519 & \textbf{0,3173} & 0,3137 &          0,2896 & 0,2790 &          0,2955 & 0,5962 & \textbf{0,2959} \\
        cond-turbine         & 0,3002 & \textbf{0,1096} & 0,1764 & \textbf{0,1094} & 0,1117 &          0,1064 & 0,1062 &          0,1094 & 0,4481 & \textbf{0,1096} \\
        cuff-less            & 0,7996 & \textbf{0,5791} & \textbf{0,5165} & 0,6255 & \textbf{0,5025} & 0,6105 & \textbf{0,5012} & 0,6225 & \textbf{0,5450} & 0,5922 \\
        electrical-grid-stab & 0,5961 & \textbf{0,3156} & 0,5545 & \textbf{0,3245} & 0,3352 & \textbf{0,3068} & \textbf{0,2634} & 0,3092 & 0,5951 & \textbf{0,3157} \\
        facebook-comment     & 0,6828 & \textbf{0,4242} & 0,4655 & \textbf{0,4429} & \textbf{0,4162} & 0,4384 & \textbf{0,4095} & 0,4397 & 0,5014 & \textbf{0,4347} \\
        geo-lat              & 0,8752 & \textbf{0,8066} & 0,9594 & \textbf{0,8538} & 0,8321 &          0,8287 & 0,8540 &          0,8353 & 0,8668 & 0,8408 \\
        greenhouse-net       & 0,6492 & \textbf{0,5418} & \textbf{0,5397} & 0,5686 & \textbf{0,5158} & 0,5638 & \textbf{0,5137} & 0,5618 & 0,6016 & \textbf{0,5599} \\
        KEGG-reaction        & 0,1803 & \textbf{0,0845} & 0,0919 & \textbf{0,0861} & 0,0837 &          0,0835 & 0,0835 &          0,0840 & 0,1172 & \textbf{0,0855} \\
        KEGG-relation        & 0,6342 & \textbf{0,1750} & 0,2454 & \textbf{0,1748} & 0,1782 & \textbf{0,1700} & 0,1809 & \textbf{0,1715} & 0,4314 & \textbf{0,1738} \\
        online-news          & 1,1822 &          1,0677 & 0,9516 & 0,9622          & \textbf{0,9177} & 0,9521 & \textbf{0,9124} & 0,9521 & \textbf{0,9228} & 0,9942 \\
        video-transcode      & 0,3975 & \textbf{0,0561} & 0,0846 & \textbf{0,0652} & 0,0604 &          0,0597 & 0,0590 & \textbf{0,0568} & 0,3704 & \textbf{0,0627} \\
        pm25-chengdu-us-post & 0,8022 & \textbf{0,4283} & 0,5578 & \textbf{0,4164} & 0,4111 &          0,4084 & 0,4125 &          0,4079 & 0,7716 & \textbf{0,4131} \\
        park-total-UPDRS     & 0,9472 & \textbf{0,7816} & 0,9416 & \textbf{0,8057} & 0,8020 &          0,7893 & 0,8196 &          0,7860 & 0,9428 & \textbf{0,8117} \\
        physico-protein      & 0,8453 & \textbf{0,5179} & 0,7676 & \textbf{0,5338} & 0,5528 & \textbf{0,5218} & 0,5744 & \textbf{0,5249} & 0,8494 & \textbf{0,5330} \\
        production-quality   & 0,4872 & \textbf{0,2807} & 0,3886 & \textbf{0,2830} & 0,2836 &          0,2779 & 0,2794 &          0,2781 & 0,3790 & \textbf{0,2832} \\
        CT-slices            & 1,9533 &          0,0504 & 0,1332 & \textbf{0,0586} & 0,0544 & \textbf{0,0418} & 0,0693 & \textbf{0,0355} & 0,1104 & \textbf{0,0376} \\
        gpu-kernel-perf      & 0,6347 & \textbf{0,0696} & \textbf{0,0300} & 0,0557 & \textbf{0,0246} & 0,0501 & \textbf{0,0228} & 0,0500 & 0,5927 & \textbf{0,0620} \\
        SML2010              & 0,2536 & \textbf{0,1157} & 0,2179 & \textbf{0,1036} & 0,1241 & \textbf{0,0857} & 0,1058 & \textbf{0,0920} & 0,3473 & \textbf{0,1221} \\
        seoul-bike-sharing   & 0,6990 & \textbf{0,4839} & 0,5801 & \textbf{0,5217} & 0,5101 &          0,5149 & 0,5171 &          0,5139 & 0,5891 & \textbf{0,5169} \\
        uber-location-price  & 0,9998 & \textbf{0,4589} & 0,6153 & \textbf{0,4789} & 0,4635 &          0,4699 & 0,4705 &          0,4688 & 0,8419 & \textbf{0,4879} \\
        year-prediction      & 0,8569 & \textbf{0,8079} & 0,8783 & \textbf{0,8137} & 0,8059 &          0,8062 & \textbf{0,7929} & 0,8131 & 0,8822 & \textbf{0,8138} \\
        \hline
        \textbf{Moyenne } & 0,7083 & 0,3842	& 0,4754 & 0,3945 & 0,3856 & 0,3842 & 0,3831 & 0,3850 & 0,5749 & 0,3917 \\
        \hline
        \multicolumn{1}{|r|}{\textbf{Déf / Egal / Vict }} & \multicolumn{2}{|c|}{ 0 / 4 / 29 }& \multicolumn{2}{|c|}{ 3 / 2 / 28 } & \multicolumn{2}{|c|}{ 8 / 16 / 9 } & \multicolumn{2}{|c|}{ 10 / 14 / 9} & \multicolumn{2}{|c|}{ 2 / 2 / 29 } \\
        \hline
        
    \end{tabular}
    \caption{Results for each dataset in Test (average results over the 10 test folds (see Section \ref{datasets})) for $S=32$.}
    \label{table:RMSEComparison}
\end{table}

The penultimate row of the table shows the average RMSE obtained on all the data sets (purely indicative), confirming the interest of the proposed method for three of the five regressors. This average should be treated with caution. As each dataset is a problem of unique difficulty, the error scale differs between each dataset. So, if the new RMSE {\it mean} in test is higher, this does not necessarily mean that the regressors perform worse on average. For this reason, in the next section we present another view of the results for further analysis.

{\bf $\bullet$ Critical diagram - } We present in Figure \ref{fig:cd} critical diagrams comparing the results obtained by each regressor through the Nemenyi post-hoc test, performed after a Friedman signed rank test of RMSE values, taking into account all the data sets.

\begin{figure}[!ht]
	\begin{center}
		\includegraphics[width=\textwidth]{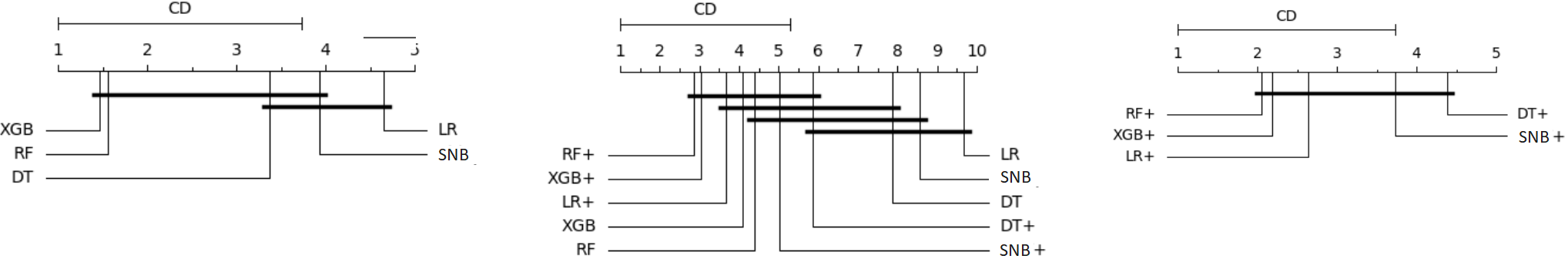}
		\caption{Critical diagram: on the left, the regressors without the proposed method, in the center, the regressors with and without the proposed method, and on the right, the regressors only with the proposed method (indicated by a '+').}
		\label{fig:cd}
	\end{center}
\end{figure}

From the graph on the left of Figure \ref{fig:cd} it can be seen that the XGBoost regressor performs best, achieving the highest mean RMSE rank, closely followed by Random Forest (RF). These two regressors outperform the other three regressors by a wide margin, with linear regression being the worst performing model. It is closely followed by the naive Bayes regressor and the regression tree.The figure \ref{fig:cd} in the middle compares the performance of all the regressors, with and without the proposed method: the first point we can observe is that, statistically, all the regressors with $X'$ have a better average rank than their counterparts with only $X$. This is an encouraging result: no regressor has been negatively affected by the proposed method, since all regressors that have used this method are better ranked than their basic versions. Again, there are different groups, and the reader can compare the change in the ranking of different regressors by comparing the figure on the left with the one in the middle.
Finally, the graph on the right of figure \ref{fig:cd} shows that: when the proposed method is used, the regressors are no longer differentiable according to the Nemenyi test (only one group compared to the left figure), even though they have different mean ranks.

\section{Conclusion}
\label{sec:Conclusion}

This paper has proposed a method for automatically generating variables  (in the case of regression) that complement the information contained in the initial input vector, the explanatory variables. The method works as a pre-processing step in which the continuous values of the variable to be regressed are discretized into a set of intervals. These intervals are used to define threshold values. Classifiers are then trained to predict whether the value to be regressed is less than or equal to each of these thresholds. The different outputs of the classifiers are then concatenated into an additional vector of variables that enriches the initial vector of explanatory variables native to the regression problem. The results are encouraging, although they mainly benefit three of the five regressors for which the method has been tested. A first improvement would be to extract a more informative vector from the classifiers, such as tree leaf identifiers. A second improvement could be to design a neural architecture that combines all the steps of the proposed method.

The proposed method also opens up certain perspectives. The attentive reader will have noticed that the vector $X'$, as defined in the proposed implementation, is in fact an estimate of the conditional cumulative density of $y$, knowing $X$. It would therefore be possible to dispense with regressors altogether, via an expectation calculation, in the case where $S$ is sufficiently high to have a fairly good estimate of this cumulative density. This last point is likely to be the subject of future work.

\bibliographystyle{splncs04}
\bibliography{biblio}

\end{document}